\documentclass{article} 
\usepackage[final]{colm2026_conference}

\usepackage[T1]{fontenc}
\usepackage{microtype}
\usepackage{amsmath}
\usepackage{amssymb}
\usepackage{graphicx}
\usepackage{booktabs}
\usepackage{array}
\usepackage{xspace}
\usepackage{hyperref}
\usepackage{url}
\usepackage{lineno}

\definecolor{darkblue}{rgb}{0, 0, 0.5}
\hypersetup{colorlinks=true, citecolor=darkblue, linkcolor=darkblue, urlcolor=darkblue}

\newcommand{\bend}{\ensuremath{B_{\text{end}}}\xspace}

\title{The Halt Vector: Internalizing a Causal Steering \\ Intervention for Efficient Reasoning}

\author{Dylan Jayabahu \\
University of Waterloo \\
\texttt{dylan.jayabahu@uwaterloo.ca} \\
\And
Tinuade Adeleke \\
Independent Researcher}

\begin{document}

\ifcolmsubmission
\linenumbers
\fi

\maketitle

\lhead{Published at the COLM 2026 Workshop on Efficient Reasoning}

\begin{abstract}
Reasoning models do not stop when they know the answer. On \texttt{DeepSeek-R1-Distill-Qwen-7B} the chain of thought runs about twice as long as the model's own answer probability takes to settle, and how much of that excess is removable varies from problem to problem, so a global length penalty cannot take it out. We take it out by internalizing a causal interpretability finding into the weights. The mechanism is a halt vector: a difference-of-means direction at layer 18 of this model whose steering strength controls how long it thinks, while a replicated value axis \citep{jiang2026} does nothing. Installing that intervention in the weights is harder than it looks. Maximizing the scalar projection onto the direction corrupts the off-axis dimensions a frozen downstream reader depends on, and generation gets longer instead of shorter; what works is reconstructing the whole steered activation with those dimensions pinned to their natural values. Fit from 24 problems and no reinforcement learning, the halt removes about a quarter of the thinking at held accuracy across five unseen benchmarks, and the cut tracks each problem's own removable slack at 0.70. It also closes a non-termination pathology that grows with difficulty and that a decoding-time confidence hook makes worse. We do not claim to beat a well-tuned length penalty \citep{arora2025,liu2025} or decoding-time early exit \citep{yang2025} on the raw trade-off; the contribution is how the halt is obtained.
\end{abstract}

\section{Introduction}\label{sec:intro}

Reasoning models answer in two visible phases: a long chain of thought delimited by \texttt{<think>} and \texttt{</think>} \citep{wei2022,guo2025,jaech2024}, then a final answer. The chain of thought improves accuracy, but these models overthink, reaching a correct answer and then re-deriving, re-checking, and occasionally reasoning back out of it \citep{chen2024,sui2025}. On \texttt{DeepSeek-R1-Distill-Qwen-7B}\footnote{Code, configs, and evaluation scripts: \url{https://github.com/dylanjayabahu/halt-vector}.} the gap between knowing and stopping is large and stable. On the base model a probe-based confidence estimate saturates at a median of about 1,650 think-tokens, and stays near there across every trained variant, while those same rollouts run to a median of about 3,700, a matched factor of 2.24. The comparison must be like for like. Thirteen percent of rollouts never saturate before the 16k cap, so they contribute no saturation point, and they are the long ones. Counting them into the numerator alone, as a median of about 4,100 emitted tokens or a mean of about 6,280 (the statistic the results tables report), inflates the ratio. The tail is net-harmful: stopping at the best internal moment would raise development-set accuracy from 0.664 to 0.823, since it reverses a correct answer on 21.0\% of problems while rescuing one on only 5.1\%.

\paragraph{Why the obvious objectives do not work.} Rewarding shortness, imitating short traces, or preferring short completions all fail to install a per-problem adaptive halt (full account in Appendix~\ref{app:negatives}). Outcome-level reinforcement learning, whether a length penalty or a per-problem budget annealed toward where the model already knows the answer, only trims the longest tails. The internal answerable point does not move, and a naive global length penalty matches or beats every per-problem budget we tried, because a single scalar reward over a 5,000-token trace cannot localize where to stop. Supervised imitation and preference methods fail complementarily: the stop-and-commit tokens are about 0.34\% of the token loss, so plain fine-tuning leaves length unchanged, length-normalized preference cancels the length signal, and preference optimization exhibits likelihood displacement; only up-weighting the stop tokens moves length, and that trades accuracy roughly one point for one. So we impose a gate before any generalization claim: a method must shorten even its own training problems at held accuracy\footnote{Throughout, ``at held accuracy'' means accuracy within cross-seed training noise, about 2 percentage points, with termination no worse than the base model. It is not a claim of provable equality; binomial sampling error at our sample sizes is about 4 points, so we lean on the grader-independent think-token and termination signals.} (Section~\ref{sec:fit}).

\paragraph{From a diagnostic to an intervention.} Our method follows from a recent interpretability result. \citet{jiang2026} construct a linear value axis from in-context reinforcement-learning data and show that steering it modulates confidence, self-correction, and verbosity. We reconstruct their difference-of-means methodology on our model and study two directions: a replication of their value axis, from an in-context guessing game, and a direction built from where the model's answer locks on during math reasoning. The value axis is near-orthogonal to this halt vector and does not govern math halting, so it serves as a control, while the halt vector is a causal stopping knob (Section~\ref{sec:causal}). This decouples the two things every failed objective conflated: how to stop, a causal actuator, and when to stop, the per-problem timing signal given by the forced-answer point.

\paragraph{Contributions.} (1) On \texttt{DeepSeek-R1-Distill-Qwen-7B}, a difference-of-means halt vector at layer 18 is a causal stopping knob: steering it shortens generation at $\rho = -0.82$, while the \citeauthor{jiang2026} value axis is flat at $+$0.09 (Section~\ref{sec:causal}). (2) Turning this into control requires reconstructing the whole steered activation with off-axis dimensions pinned; a scalar-projection target instead corrupts off-axis dimensions and lengthens generation (Section~\ref{sec:internalize}). (3) The result is a hook-free, per-problem self-halt, fit from 24 problems without reinforcement learning: about 24\% less thinking at held accuracy across five unseen benchmarks, per-problem adaptive at a slack-to-cut correlation of 0.70, stable over three seeds (Section~\ref{sec:results}). (4) A magnitude-matched direction-specificity control, value axis or random direction, installs no halt (Section~\ref{sec:dirspecific}). (5) The halt also fixes base non-termination, a pathology that grows with difficulty (Section~\ref{sec:termination}). (6) Scaling and failure modes: single-run training erodes the halt, a weight-space burst-merge scales the data safely, truncation is a clean negative, and our same-grader reimplementation of the strongest decoding-time competitor holds accuracy where that competitor trades it away on the hard sets (Sections~\ref{sec:scaling} and~\ref{sec:comparison}).

\paragraph{What we do not claim.} We do not beat a well-tuned length-penalty reinforcement-learning method \citep{arora2025,liu2025} or a decoding-time early-exit method \citep{yang2025} on the raw compression-accuracy trade-off. Our efficiency is competitive rather than state of the art; the contribution is the mechanism and a deployment profile with no serving-time overhead and no reinforcement-learning pipeline.

\section{Related work}\label{sec:related}

\paragraph{Efficient reasoning.} Overthinking in reasoning models is well documented \citep{chen2024,sui2025}, and mitigations cluster into four groups: length-penalty reinforcement learning, which rewards shorter correct rollouts \citep{arora2025,luo2025,aggarwal2025}, with DLER \citep{liu2025} the state of the art; test-time budget control, which caps the token budget at inference \citep{muennighoff2025}; decoding-time dynamic early exit, which stops once a confidence signal crosses a threshold \citep{yang2025}; and supervised or prompt-based compression of the chain of thought \citep{xu2025}. DLER is the closest, converging on two ideas adjacent to ours, an update-selective weight merge and difficulty-aware truncation, yet it remains outcome-level reinforcement learning over many problems. Our halt is orthogonal to all four: it is per-problem and learned from a causal internal signal.

\paragraph{Reading reasoning state.} A parallel line shows hidden states linearly encode whether a chain of thought will succeed before it completes \citep{afzal2025} and whether an intermediate answer is correct \citep{zhang2025}. \citet{zhang2025} use such a probe as a verifier to trigger an early exit, reporting a roughly 24\% token reduction at no accuracy cost. Our target is a ground-truth forced-answer measurement rather than a learned probe, and we install the halt in the weights, so inference needs no auxiliary model.

\paragraph{Activation steering and internalizing directions.} A large body of work constructs concept directions as a difference of means and reads or steers them at inference \citep{li2023,turner2023,rimsky2024,arditi2024}, part of the broader representation-engineering program \citep{zou2023}. Closest to us is work that moves a direction into the weights: \citet{arditi2024} bake in a refusal ablation by weight orthogonalization, \citet{ackerman2024} fine-tunes a steering vector into the residual stream by maximizing cosine similarity, and \citet{fierro2025} isolate a behavior direction in weight space and argue such edits generalize better than activation steering. We differ in two ways that matter for the mechanism: prior representation tuning maximizes a scalar cosine similarity, which we show fails here because it corrupts off-axis dimensions, so reconstructing the whole steered activation is required; and we freeze the downstream reader and target a per-problem point, installing a conditional halt rather than a global behavior shift.

\paragraph{The value axis.} \citet{jiang2026} construct a value axis from in-context reinforcement learning and show that steering it modulates verbalized confidence, backtracking, and verbosity on Qwen3-8B. We adopt their difference-of-means construction (Equations~\ref{eq:diffmeans} to~\ref{eq:steer}), but on \texttt{DeepSeek-R1-Distill-Qwen-7B} their value axis is near-orthogonal to our halt vector and does not govern math halting, so it is our control (Sections~\ref{sec:causal} and~\ref{sec:dirspecific}). Concurrent work finds that reinforcement learning recruits similar functional axes \citep{han2026} and that verbal uncertainty is a manipulable linear feature \citep{ji2025}.

\paragraph{Model merging.} Our safe-scaling result builds on weight-space merging: we average independently-trained LoRA adapters \citep{hu2022} in the manner of model soups \citep{wortsman2022} and compare against the sign-aware TIES recipe \citep{yadav2023}, both understood through task arithmetic, in which a fine-tuned model minus its base defines a composable task vector \citep{ilharco2023}.

\section{Background: measuring when the model knows}\label{sec:background}

\paragraph{Setup.} All training and evaluation ran on Modal (H200 GPUs). The model throughout is \texttt{DeepSeek-R1-Distill-Qwen-7B} \citep{guo2025}, 7B parameters, 28 layers numbered 0 to 27. Training problems and offline pools are drawn from DeepScaleR \citep{deepscaler2025}; we evaluate zero-shot on the full MATH500 \citep{hendrycks2021}, AMC \citep{amc}, and AIME 2024/2025 \citep{aime} sets. We reserve \emph{value axis} for the direction of \citet{jiang2026}, built from in-context reinforcement-learning data and a control on our model (Section~\ref{sec:causal}), and call our own difference-of-means direction, built from math reasoning positions before and after the model's answer locks on, the \textbf{halt vector}.

\paragraph{The forced-answer point.} Training a model to stop when it knows needs a trustworthy measurement of when it knows, and a probe-based confidence estimate is not safe to cut on: a probe checking only the first token of the gold answer often fires on the answer-format prior before any reasoning, and reconstructing what it has seen at high confidence shows neither the gold nor the model's eventual answer on the page in 93 to 100\% of easy-pool cases. We therefore measure ground truth: we take the reasoning up to some point, append a short scaffold \verb|\n</think>\n\n**Final answer:** $\boxed{|, let the model greedily emit the answer, and grade it (\texttt{math\_verify}, which treats $1/2$, $0.5$, and $\frac{1}{2}$ as equal). The forced-answer point \bend is the earliest prefix at which this forced answer is already correct; it is free-running and format-robust, and everything after it is in principle cuttable slack. A problem is \emph{compressible} (\emph{goldilocks}) when \bend leaves substantial slack and an \emph{anchor} when it has none; the 24-problem training set and its disjoint holdout are compressible.

Measured offline on the base model and requiring the gold answer, \bend is a training target, not a deployable signal; turning ``can be forced to answer here'' into ``chooses to stop here'' is the object of this work.

\paragraph{Two directions.} Following \citet{jiang2026}, we construct difference-of-means directions on our model. For a contrast set $C$ in which each item has a set of ``before'' positions $T_{\text{pre}}$ and a set of ``after'' positions $T_{\text{post}}$, the per-layer direction is
\begin{equation}\label{eq:diffmeans}
a^{(\ell)} = \frac{1}{|C|}\sum_{c \in C}\Big(\tfrac{1}{|T^c_{\text{post}}|}\!\!\sum_{t\in T^c_{\text{post}}}\!\! h^{(\ell)}_{c,t} - \tfrac{1}{|T^c_{\text{pre}}|}\!\!\sum_{t\in T^c_{\text{pre}}}\!\! h^{(\ell)}_{c,t}\Big), \qquad \hat u^{(\ell)} = a^{(\ell)} / \lVert a^{(\ell)}\rVert,
\end{equation}
where $h^{(\ell)}_{c,t}$ is the residual-stream activation at layer $\ell$ for item $c$ at token position $t$, and $a^{(\ell)}$ is the resulting unnormalized difference-of-means direction. We read it out by cosine projection, so that a token's projection is
\begin{equation}\label{eq:proj}
\operatorname{proj}(h_t) = \hat u^{(\ell)} \cdot h_t / \lVert h_t\rVert.
\end{equation}
We build two directions, both positional contrasts of the form in Equation~\ref{eq:diffmeans}: a replication of the \citeauthor{jiang2026} value axis from an in-context guessing-game contrast (AUROC 0.78 at layer 18, held out by conversation with disjoint criteria), and the halt vector, which contrasts think-token positions after the point at which the model's probability on the first token of the gold answer reaches 0.9 and stays there against positions before it, within rollouts the model answers correctly (AUROC 0.76 at layer 18 for that positional discrimination, held out by rollout rather than by problem; see Appendix~\ref{app:axis}). Two disclosures. This is a within-rollout lock-on contrast, not a between-item contrast of correct against incorrect reasoning; we built that variant and discarded it as surface-confounded, near 0.96 but flat from layer 0. And the first-token probability is the same measure of knowing that Appendix~\ref{app:outcomerl} rejects as an artifact; it places the contrast used to build the direction, is never a training target, and the direction's value rests on the causal test of Section~\ref{sec:causal} rather than on this AUROC (Appendix~\ref{app:axis} gives the construction in full). The two are near-orthogonal (cosine similarity near zero), so guessing-game confidence and this lock-on feature are distinct internal features here. Neither is a reliable answerability gauge, since the internal projection keeps rising well past \bend, so internal belief precedes verifiable answerability. The halt vector is also the weaker read-out. At layer 18 the value axis separates slightly better, on the cleaner split. What distinguishes the halt vector is actuation, not separability, which is what Section~\ref{sec:causal} tests. That a direction built from the lock-on point should also be a stopping actuator is unsurprising: it separates late reasoning positions from early ones, so steering toward the late end pushes the model toward what it does at the end of a trace, which is commit and stop. Whether stopping there is safe is a separate question, and \bend is what answers it.

\section{The halt vector is a causal stopping knob}\label{sec:causal}

We now test whether steering the halt vector moves the point at which the model stops. During generation we add $(\alpha/100)\,\lVert\bar h\rVert\,\hat u$ to the residual stream at layer $\ell$, where $\alpha$ is a percentage of the mean activation norm $\lVert\bar h\rVert$, and we sweep $\alpha$:
\begin{equation}\label{eq:steer}
h^{(\ell)}_t \leftarrow h^{(\ell)}_t + \tfrac{\alpha}{100}\,\lVert\bar h\rVert\,\hat u^{(\ell)}.
\end{equation}
The result, at layer 6, is a clean monotonic knob (Table~\ref{tab:layer6}, in the appendix).

\paragraph{The halt vector is a causal length knob.} Steering strength and length correlate at $-$0.60 at layer 6 on the 128-problem knob demo of Table~\ref{tab:layer6} (the 64-problem layer sweep below independently measures $-$0.63): positive steering makes the model wrap up almost immediately, negative steering prevents termination entirely. A magnitude-matched value-axis control applied the same way shows no length effect ($\rho = +0.09$), so the effect is specific to the halt vector, not to perturbing the residual stream by a comparable amount in any direction (Section~\ref{sec:dirspecific} corroborates this at layer 18 on the training side). Two facts shape the method. First, the accuracy collapse under steering is acceptable, because the knob's only role is to be a reliable stopping actuator and correctness comes from where it fires: applied at or after \bend, where the forced answer is correct by construction, the model stops and is right. Second, a sweep over layers 6, 12, 18, and 24 shows later layers preserve accuracy far better than early ones, so the layer-6 collapse is an early-layer artifact, and layer 18 is the best knob.

\paragraph{Layer sweep.} Steering the halt vector at four layers (64 problems, strength $-$50 to $+$50; base at strength 0: 4,746 median think-tokens, 80\% terminated, 72\% accuracy; Table~\ref{tab:layersweep}, in the appendix), retained accuracy at $+$10 climbs from 14\% at layer 6 to 56\% at layer 18 to 72\% at layer 24. But layer 18 gives the strongest, cleanest length control ($\rho=-0.82$; $+$10 yields 31\% shorter generation and more frequent termination at accuracy near 56\%), while layer 24 holds accuracy best but is too weak a lever ($-$0.24). All internalization below is therefore at layer 18.

\section{Internalizing the knob}\label{sec:internalize}

We split the network at layer 18 into a writer (layers 0 to 18), which produces the layer-18 activation, and a reader (layers 19 to 27), which interprets it. Placing LoRA only on the writer freezes the reader, preventing the direction's meaning from drifting, and we train the writer to make its own layer-18 activation match the steered one after \bend.

\paragraph{A scalar-projection target fails.} The natural objective is to make the scalar projection onto $\hat u$ large after \bend. We swept it hard, with larger targets, an unbounded variant, and added MLP capacity, and it fails even in-distribution. The harder we push the projection, the longer and worse the model becomes. The strongest arm reached a projection of $+$0.34 but generated 43\% longer, fell to 0.57 accuracy, and stopped emitting \texttt{</think>} altogether. Satisfying a single scalar leaves the off-axis dimensions unconstrained, and they move. Measured teacher-forced at layer 18, the reconstruction arms reach at most 3.66 units of off-axis drift per unit of achieved on-axis movement against the scalar arms' minimum of 5.22, with no overlap between the families on raw drift, on that normalized ratio, on the fraction of tokens leaving the base model's own p99 off-axis band, or on the median reader KL; and drift ranks against behavioural damage at $\rho = -0.799$ (permutation $p = 0.008$). The drift nonetheless stays inside the base model's own off-axis dispersion, so the frozen reader is disturbed rather than handed a grossly out-of-distribution vector, and the model rambles and loops (Figure~\ref{fig:mechanism}; full table in Appendix~\ref{app:drift}). This is also why prior representation tuning that maximizes cosine similarity to a direction \citep{ackerman2024} is insufficient here: cosine similarity is a scalar and leaves the writer free to distort the off-axis dimensions. The converse repair, adding an off-axis preservation penalty to the scalar objective, does not answer whether off-axis freedom is what breaks the scalar target. At every weight we tried it suppressed the edit rather than cleaning it, driving achieved on-axis movement to near zero along with the drift, so the arm never reached a state where the comparison could be made (Appendix~\ref{app:perp}).

\paragraph{Reconstructing the whole steered activation works.} Instead of a scalar, we regress the entire layer-18 vector to what steering would have produced. For a token at position $t$ the target is
\begin{equation}\label{eq:recon}
h^{*}(t) = \underbrace{h_{\text{base}}(t)}_{\text{off-axis pinned to natural}} + s(t)\,\tfrac{\alpha}{100}\,\lVert\bar h\rVert\,\hat u,
\end{equation}
where $h_{\text{base}}(t)$ is the base activation at position $t$, obtained with the adapter disabled, $\lVert\bar h\rVert$ is the base model's mean layer-18 think-token norm as a fixed scalar reference, $\alpha$ is the steering strength, and $s(t)$ is a schedule that reaches its peak at \bend. The training loss combines a per-position reconstruction error over the think tokens with the ordinary language-model loss on the full trace:
\begin{equation}\label{eq:loss}
\mathcal{L} = \mathcal{L}_{\text{LM}} + \lambda\cdot\frac{1}{|T_{\text{think}}|}\sum_{t\in T_{\text{think}}} \frac{\lVert h(t) - h^{*}(t)\rVert^2}{(\tfrac{\alpha}{100}\lVert\bar h\rVert)^2}.
\end{equation}
Because the target sets only the on-axis component and copies $h_{\text{base}}$ elsewhere, it pins the off-axis dimensions to their natural values and hands the frozen reader exactly the vector it responds to (Figure~\ref{fig:mechanism}).

\paragraph{Ablations.} Three ablations pin down the mechanism. First, the halt needs the sustained steer: holding it after \bend installs the halt, while raising then releasing it does no better than base. Second, the pre-\bend ramp shape is irrelevant, since a gated schedule (zero before \bend, one after) matches the held one. Third, attention-only adaptation beats adding the MLP, because extra capacity gives the writer more freedom to corrupt off-axis dimensions. The headline adapter therefore reconstructs the halt vector with a held schedule, steering strength 25, attention-only LoRA on layers 0 to 18, loss weight 1, seed 17, and 12 epochs on 24 problems (details in Appendix~\ref{app:impl}).

\begin{figure}[t]
\begin{center}
\includegraphics[width=0.9\linewidth]{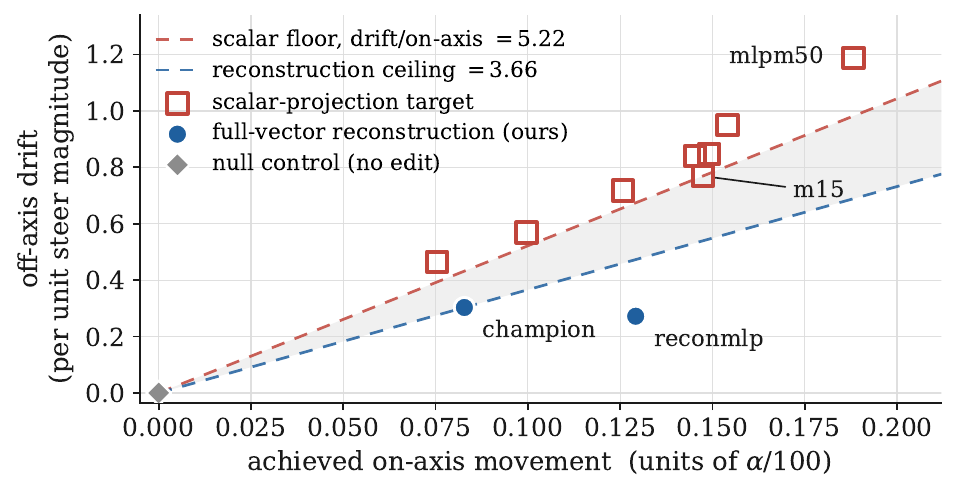}
\end{center}
\caption{\textbf{Off-axis drift separates the two objectives.} Achieved on-axis movement against off-axis drift at layer 18, both in units of the injected steer magnitude $(\alpha/100)\lVert\bar h\rVert$, measured teacher-forced on think tokens with the adapter enabled and disabled. Each point is one trained arm. The families do not overlap, and the shaded wedge between the two ratio bounds is empty: every scalar arm drifts further off-axis than every reconstruction arm, and the gap only widens once drift is normalized by the on-axis movement actually achieved. Drift stays within the base model's own off-axis dispersion throughout, so this is comparative disturbance of the frozen reader, not a grossly out-of-distribution activation. Full table in Appendix~\ref{app:drift}.}
\label{fig:mechanism}
\end{figure}

\section{Results}\label{sec:results}

\subsection{Fitting the halt on 24 problems}\label{sec:fit}
No prior objective produced a result both shorter and accurate even on its training problems (Appendix~\ref{app:negatives}), so we first confirm ours can. We fit 24 compressible problems with a LoRA adapter and no reinforcement learning; the base model scores 0.812 accuracy at 6,516 think-tokens with 0.95 termination (Table~\ref{tab:indist}, in the appendix).

Only reconstruction compresses at held accuracy, and it terminates more cleanly than base: it cuts think-tokens 25\% at 0.792 accuracy ($-$2.0 points), whereas imitation runs 15\% longer ($-$10.4 points) and gated steering-distillation cuts 21\% only by trading away 6.2 points. It is the first objective in our study to weight the stopping decision without sacrificing accuracy.

\paragraph{A tunable dial.} Sweeping strength over 10, 25, 40, and 50 yields think reductions of 11\%, 25\%, 27\%, and 31\% while point-estimate accuracy declines monotonically over 0.833, 0.792, 0.781, and 0.740. We take 25 as the headline operating point on two grounds: compression saturates there (40 adds only 2 further points, 50 adds 6, while accuracy slides, so 40 is weakly dominated), and 25 is the most aggressive setting whose accuracy stays within about 2 points of base, roughly the 2.6-point cross-seed standard deviation over our three seeds. These differences are not statistically resolvable at our sample sizes, so 25 is conservative, not a claim that stronger settings hurt, and there is no accuracy knee. The gentle end, strength 10, leaves the held-out anchor problems essentially untouched, so the two settings span selective to aggressive.

\paragraph{The halt is per-problem.} To check that the halt conditions on the problem, we evaluate held-out anchor problems, which have no early answer point and on which base scores 0.719 accuracy at 4,812 think-tokens. At strength 10 their length is essentially unchanged ($+$2\%) and their accuracy exact, while the compressible problems are cut 11\%, so the halt fires only when the model actually reaches an answerable state early. Higher strengths trim the anchors too, but anchor accuracy never declines across the sweep (Table~\ref{tab:anchorsplit}, in the appendix), so those cuts are probe-missed slack rather than eaten reasoning. The anchor grader is noisier (comparator agreement 0.41 to 0.53), so there we lean on length and the no-collapse trend.

\subsection{The halt is direction-specific}\label{sec:dirspecific}
A skeptic might object that reconstructing any layer-18 target at \bend installs a halt, making the effect about the landmark rather than the direction. We re-run the headline recipe changing only $\hat u$; the injected magnitude $(\alpha/100)\lVert\bar h\rVert$ is the same for every direction, so magnitude is fixed. The controls are the value axis, a real but non-halting feature, and an isotropic random direction.

\begin{table}[t]
\centering
\small
\setlength{\tabcolsep}{5pt}
\begin{tabular}{lrrrrrr}
\toprule
 & \multicolumn{3}{c}{train-24} & \multicolumn{3}{c}{AMC} \\
\cmidrule(lr){2-4}\cmidrule(lr){5-7}
direction distilled at \bend & acc & think & $\Delta$ & acc & think & $\Delta$ \\
\midrule
\textbf{halt vector (ours)} & 0.792 & 4911 & $-$25\% & 0.822 & 4760 & $-$24\% \\
value axis (control) & 0.771 & 6253 & $-$4\% & 0.797 & 6162 & $-$2\% \\
random direction (control) & 0.771 & 7195 & $+$10\% & 0.807 & 6468 & $+$3\% \\
\bottomrule
\end{tabular}
\caption{\textbf{Direction specificity at fixed magnitude.} The headline recipe re-run with only the distilled direction $\hat u$ changed. With magnitude held fixed, only the halt vector installs the halt.}
\label{tab:dirspecific}
\end{table}

The value axis barely moves generation and the random direction runs longer, the same off-axis-corruption signature as the failed scalar target (Table~\ref{tab:dirspecific}). This is the training-side counterpart of the flat value-axis steering control in Section~\ref{sec:causal}. The value axis alone might be dismissed as off-domain, but the random arm is equally magnitude-matched and still fails.

\subsection{Generalization}\label{sec:generalization}
We now evaluate the 24-problem adapter zero-shot, at a 16k-token cap, on five held-out benchmarks (Table~\ref{tab:generalization}).

\begin{table}[t]
\centering
\small
\begin{tabular}{lllrr}
\toprule
eval set & base (acc / think) & ours (acc / think) & $\Delta$think & closed \\
\midrule
goldilocks holdout (disjoint) & 0.845 / 6617 & 0.835 / 5014 & $-$24\% & 0.98 \\
AMC (cross-benchmark) & 0.813 / 6280 & 0.822 / 4760 & $-$24\% & 0.98 \\
AIME24 & 0.512 / 10413 & 0.492 / 7898 & $-$24\% & 0.94 \\
AIME25 & 0.358 / 10808 & 0.385 / 8376 & $-$22\% & 0.91 \\
MATH500 \citep{hendrycks2021} & 0.934 / 3101 & 0.930 / 2495 & $-$20\% & 0.99 \\
\bottomrule
\end{tabular}
\caption{\textbf{Held-out generalization.} The 24-problem adapter evaluated zero-shot at a 16k-token cap on five held-out benchmarks; the goldilocks holdout is 100 disjoint compressible problems. Reconstruction removes 20 to 24\% of thinking at held accuracy throughout, and the dial carries over ($\alpha=10$ gives 11 to 16\%). The goldilocks-holdout row is graded at $n=4$; the matched $n=8$ read is 0.819 accuracy at $-$22.0\% (Table~\ref{tab:burst}).}
\label{tab:generalization}
\end{table}

The cut holds even on the easy MATH500, already short at about 3,100 think-tokens, so it removes slack across difficulty rather than only taming long AIME traces. With the direction-specificity control, this rules out overfitting to the 24 training problems.

\begin{figure}[t]
\begin{center}
\includegraphics[width=\linewidth]{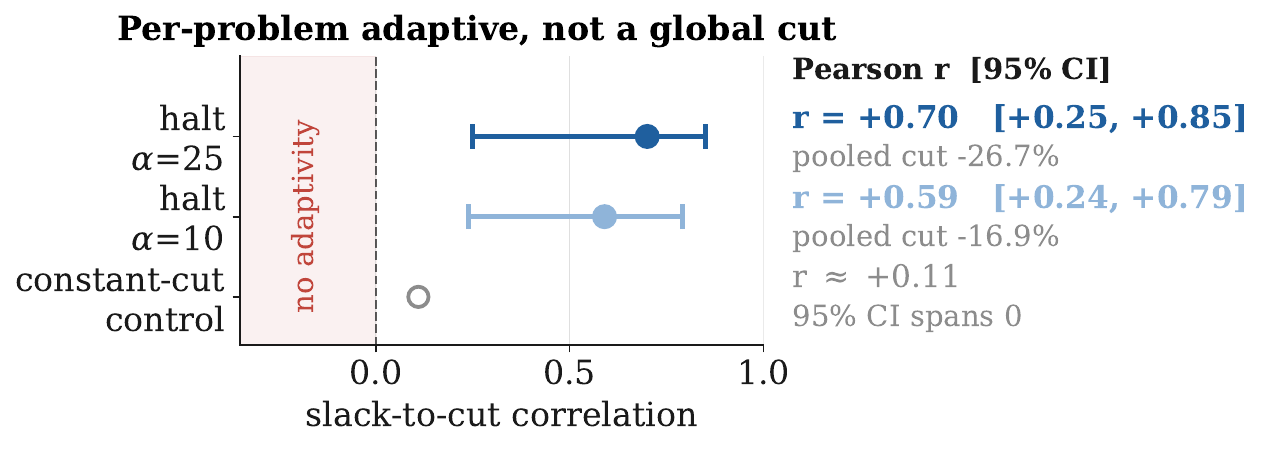}
\end{center}
\caption{\textbf{Per-problem adaptivity.} Slack-to-cut Pearson correlation with bootstrap 95\% confidence intervals for the two halt strengths and the constant-cut control. Both halt intervals sit entirely above zero while the control straddles it. The figure reports point estimates and intervals; the individual points' raw values reside on the run volume.}
\label{fig:adaptivity}
\end{figure}

\paragraph{Per-problem adaptive at scale.} To quantify adaptivity, we correlate each of 100 holdout problems' realized think-cut with its own compressible slack, $1 - B_{\text{end}}/\text{full}$, a reference the model never sees. The Pearson coefficient is 0.70 (95\% CI 0.25 to 0.85) at strength 25 and 0.59 (CI 0.24 to 0.79) at strength 10, both excluding zero, while a synthetic constant-cut control scores 0.11 with an interval spanning zero (Figure~\ref{fig:adaptivity}). The halt cuts more where a problem has more cuttable tail and less where it does not, which separates it from a global length penalty.

\subsection{The halt fixes a termination pathology}\label{sec:termination}
\begin{figure}[t]
\begin{center}
\includegraphics[width=0.834\linewidth]{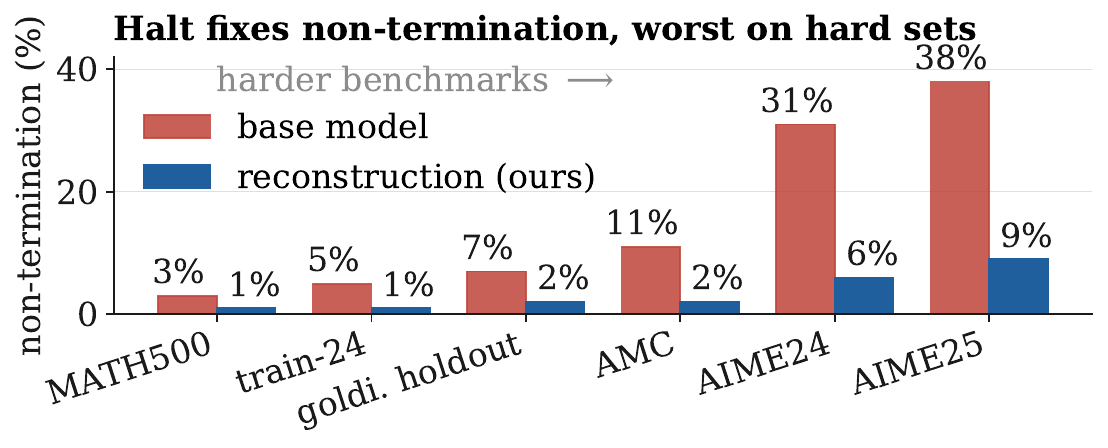}
\end{center}
\caption{\textbf{Closure by difficulty.} The figure shows the base and reconstruction non-termination rates side by side, ordered by difficulty.}
\label{fig:closure}
\end{figure}

Beyond trimming completed answers, the halt addresses a separate failure mode: the base model often runs to the 16k-token cap without terminating, at a rate that grows with difficulty, and reconstruction closes this gap on every benchmark (Figure~\ref{fig:closure}; full numbers in Table~\ref{tab:termination}).

On the two AIME sets non-termination falls from 31 to 38\% down to 6 to 9\%. This is a different axis from the token cut, since the model reaches its answer and stops rather than hitting the cap, and the failure mode a decoding-time confidence hook worsens on hard problems (Section~\ref{sec:comparison}); because the non-terminating tail dominates worst-case latency and cost, it is a practical gain that mean length hides.

\subsection{Seed robustness}\label{sec:seeds}
Two further replicates change only the seed. Averaged over the three, think reduction is $25.0\% \pm 2.3$ points on train-24 and $25.1\% \pm 1.0$ on AMC, termination at or above 0.97 for every seed; the AMC spread is 1.0 points across seeds (Table~\ref{tab:perseed}).

\subsection{The efficiency frontier}\label{sec:frontier}
Finally we place the halt on the compression-accuracy trade-off against our in-house DEER replication and a published length-penalty method (Table~\ref{tab:frontier}). We hold accuracy where the others trade it. On AIME24, at nearly equal compression ($-$24 against $-$26\%), we lose 2.0 points against DEER's 6.7.

\begin{table}[t]
\centering
\begin{tabular}{lrrrr}
\toprule
 & \multicolumn{2}{c}{reconstruction (ours)} & \multicolumn{2}{c}{DEER} \\
\cmidrule(lr){2-3}\cmidrule(lr){4-5}
benchmark & $\Delta$think & $\Delta$acc & $\Delta$think & $\Delta$acc \\
\midrule
goldilocks holdout & $-$24\% & $-$1.0 & $-$64\% & $+$5.0 \\
AMC & $-$24\% & $+$0.9 & $-$40\% & $-$8.4 \\
AIME24 & $-$24\% & $-$2.0 & $-$26\% & $-$6.7 \\
AIME25 & $-$22\% & $+$2.7 & n/a & n/a \\
MATH500 & $-$20\% & $-$0.4 & $-$54\% & $-$4.0 \\
\bottomrule
\end{tabular}
\caption{\textbf{Efficiency frontier: reconstruction versus DEER, each against its own base.} Think-token reduction and accuracy change (percentage points). Ours is sampled ($n \ge 4$) against its own base; DEER \citep{yang2025} is greedy ($n=1$) against a matched never-exit anchor (per-threshold cells in Appendix~\ref{app:deer}). Ours loses at most 2 points on every benchmark while cutting 20 to 24\%; DEER compresses more but trades accuracy on the hard sets. A published length-penalty RL reference \citep{arora2025} reaches $-$2.2 points at 36\% on MATH and $-$4.0 at 27\% on AIME24 on its own base.}
\label{tab:frontier}
\end{table}

\section{Scaling and robustness}\label{sec:scaling}

More single-run training \emph{erodes} the halt (150 problems at 12 epochs cut only 0.2\% against 24 problems' 24\%), because the full-trace language-model loss out-competes the reconstruction target with exposure; truncating the traces to counter this is a clean negative, buying length only by spending accuracy (Appendix~\ref{app:truncation}). Since erosion is a single-run effect, we instead average independently-trained 24-problem LoRA bursts in weight space (Table~\ref{tab:burst}): the cut holds flat at 24 to 26\% at held accuracy from 24 up to 144 problems with termination at or above 0.97, so more data is usable safely, as robustness rather than a scaling law, and a TIES-merge check rules out sign cancellation in the linear merge.

\section{Comparison to prior methods}\label{sec:comparison}

We compare against the strongest decoding-time competitor, DEER \citep{yang2025}, a training-free hook that forces \texttt{</think>} once hesitation-token confidence exceeds a threshold; we reimplement it under our grader (greedy, $n=1$, with a never-exit anchor for a fair delta), and its $\lambda=0.95$ points are the DEER column of the frontier (Table~\ref{tab:frontier}), with per-threshold cells in Appendix~\ref{app:deer}. The greedy base runs to the cap 23 to 53\% of the time on the hard and goldilocks sets, so DEER partly rescues greedy decoding from its own loops; it improves accuracy only on the goldilocks holdout and trades it elsewhere, notably by 6.7 points on hard AIME24, reproducing its own report that confidence thresholds degrade on hard problems. The two mechanisms are largely complementary: stacked on our adapter DEER is accuracy-safe on AMC, the holdout, and AIME24 (gains of 1.2/3.0/0.0 points at cuts of 33 to 66\%), though it over-cuts the easy MATH500 by 5.8 points.

DEER is a strong compressor but a different category, run-time machinery, so we do not claim to beat it on raw compression; its accuracy dip on hard sets is exactly the failure mode a learned-in halt avoids. The same holds among trained methods, where a well-tuned length penalty is at least as efficient on the raw trade-off \citep{liu2025,arora2025}. Our claim is on mechanism and deployment.

\section{Discussion and limitations}\label{sec:discussion}

\paragraph{What the mechanism means.} This turns an interpretability finding into an installed behavior: a direction found by read-out is shown causal, and the intervention is moved into the weights so it fires with no run-time machinery.

\paragraph{Limitations.} Our study is confined to a single model, \texttt{DeepSeek-R1-Distill-Qwen-7B}; extending to other families and sizes is the most important next step. The 24-problem fit is data efficiency de-risked by burst-merge, but leaves training-set accuracy noisy, so we lean on think-tokens and the larger held-out sets. The forced-answer point is measured offline and needs the gold answer, so although the deployed artifact is hook-free, the training pipeline still needs a base-model probe. Finally, accuracy is grader-sensitive: we report \texttt{math\_verify} with comparator agreement and bootstrap intervals, and the DEER comparison is a landscape, not an identical-decoding head-to-head ($n=8$ versus greedy $n=1$).

\paragraph{Future work.} Beyond other model families and sizes, further directions are a reader-side mirror (freeze the writer, train the reader on the natural direction) to localize where the competence to stop lives, and closing the training loop by re-deriving \bend on the trained model (gated on correctness); we would also test outside mathematics and make the steering strength itself per-problem.

\section{Conclusion}\label{sec:conclusion}

Reasoning models overthink by a wide, stable margin. We internalize a causal stopping direction, the halt vector at layer 18, into the weights, hook-free and without reinforcement learning, yielding a tunable, per-problem self-halt from 24 problems that removes about 24\% of thinking at held accuracy across five unseen benchmarks, tracks a hidden per-problem reference, and closes a difficulty-growing termination pathology, a competitive rather than dominant point on the frontier. Beyond halting, full-vector reconstruction with off-axis dimensions pinned is a general recipe for internalizing steering interventions, whose central failure mode, off-axis corruption, is the part most likely to recur when the technique is reused.

\bibliography{colm2026_conference}

\begin{thebibliography}{32}
\providecommand{\natexlab}[1]{#1}
\providecommand{\url}[1]{\texttt{#1}}
\expandafter\ifx\csname urlstyle\endcsname\relax
  \providecommand{\doi}[1]{doi: #1}\else
  \providecommand{\doi}{doi: \begingroup \urlstyle{rm}\Url}\fi

\bibitem[Ackerman(2024)]{ackerman2024}
C.~Ackerman.
\newblock {Representation Tuning}.
\newblock \emph{arXiv preprint arXiv:2409.06927}, 2024.

\bibitem[Afzal et~al.(2025)Afzal, Matthes, Chechik, and Ziser]{afzal2025}
A.~Afzal, F.~Matthes, G.~Chechik, and Y.~Ziser.
\newblock {Knowing Before Saying: LLM Representations Encode Information About
  Chain-of-Thought Success Before Completion}.
\newblock In \emph{Findings of the Association for Computational Linguistics:
  ACL}, 2025.

\bibitem[Aggarwal \& Welleck(2025)Aggarwal and Welleck]{aggarwal2025}
P.~Aggarwal and S.~Welleck.
\newblock {L1: Controlling How Long a Reasoning Model Thinks with Reinforcement
  Learning}.
\newblock In \emph{Conference on Language Modeling (COLM)}, 2025.
\newblock arXiv:2503.04697.

\bibitem[Arditi et~al.(2024)Arditi, Obeso, Syed, Paleka, Panickssery, Gurnee,
  and Nanda]{arditi2024}
A.~Arditi, O.~Obeso, A.~Syed, D.~Paleka, N.~Panickssery, W.~Gurnee, and
  N.~Nanda.
\newblock {Refusal in Language Models Is Mediated by a Single Direction}.
\newblock In \emph{Advances in Neural Information Processing Systems
  (NeurIPS)}, 2024.
\newblock arXiv:2406.11717.

\bibitem[Arora \& Zanette(2025)Arora and Zanette]{arora2025}
D.~Arora and A.~Zanette.
\newblock {Training Language Models to Reason Efficiently}.
\newblock In \emph{Advances in Neural Information Processing Systems
  (NeurIPS)}, 2025.
\newblock arXiv:2502.04463.

\bibitem[Chen et~al.(2024)]{chen2024}
X.~Chen et~al.
\newblock {Do NOT Think That Much for 2+3=? On the Overthinking of o1-Like
  LLMs}.
\newblock \emph{arXiv preprint arXiv:2412.21187}, 2024.

\bibitem[Fierro \& Roger(2025)Fierro and Roger]{fierro2025}
C.~Fierro and F.~Roger.
\newblock {Steering Language Models with Weight Arithmetic}.
\newblock \emph{arXiv preprint arXiv:2511.05408}, 2025.

\bibitem[Guo et~al.(2025)]{guo2025}
D.~Guo et~al.
\newblock {DeepSeek-R1 incentivizes reasoning in LLMs through reinforcement
  learning}.
\newblock \emph{Nature}, 645:\penalty0 633--638, 2025.
\newblock arXiv:2501.12948.

\bibitem[Han et~al.(2026)Han, Chalmers, and Izmailov]{han2026}
A.~Q. Han, D.~J. Chalmers, and P.~Izmailov.
\newblock {How's It Going? Reinforcement Learning in Language Models Recruits a
  Functional Welfare Axis}.
\newblock \emph{arXiv preprint arXiv:2605.30232}, 2026.

\bibitem[Hendrycks et~al.(2021)]{hendrycks2021}
D.~Hendrycks et~al.
\newblock {Measuring Mathematical Problem Solving with the MATH Dataset}.
\newblock In \emph{NeurIPS Datasets and Benchmarks Track}, 2021.
\newblock arXiv:2103.03874.

\bibitem[Hu et~al.(2022)]{hu2022}
E.~J. Hu et~al.
\newblock {LoRA: Low-Rank Adaptation of Large Language Models}.
\newblock In \emph{International Conference on Learning Representations
  (ICLR)}, 2022.
\newblock arXiv:2106.09685.

\bibitem[Ilharco et~al.(2023)]{ilharco2023}
G.~Ilharco et~al.
\newblock {Editing Models with Task Arithmetic}.
\newblock In \emph{International Conference on Learning Representations
  (ICLR)}, 2023.
\newblock arXiv:2212.04089.

\bibitem[Jaech et~al.(2024)]{jaech2024}
A.~Jaech et~al.
\newblock {OpenAI o1 System Card}.
\newblock \emph{arXiv preprint arXiv:2412.16720}, 2024.

\bibitem[Ji et~al.(2025)]{ji2025}
Z.~Ji et~al.
\newblock {Calibrating Verbal Uncertainty as a Linear Feature to Reduce
  Hallucinations}.
\newblock \emph{arXiv preprint arXiv:2503.14477}, 2025.

\bibitem[Jiang et~al.(2026)Jiang, Kauvar, and Lindsey]{jiang2026}
N.~Jiang, I.~Kauvar, and J.~Lindsey.
\newblock {The Value Axis: Language Models Encode Whether They're on the Right
  Track}.
\newblock \emph{arXiv preprint arXiv:2606.17056}, 2026.

\bibitem[Li et~al.(2023)Li, Patel, Vi\'{e}gas, Pfister, and Wattenberg]{li2023}
K.~Li, O.~Patel, F.~Vi\'{e}gas, H.~Pfister, and M.~Wattenberg.
\newblock {Inference-Time Intervention: Eliciting Truthful Answers from a
  Language Model}.
\newblock In \emph{Advances in Neural Information Processing Systems
  (NeurIPS)}, 2023.
\newblock arXiv:2306.03341.

\bibitem[Liu et~al.(2025)]{liu2025}
S.-Y. Liu et~al.
\newblock {DLER: Doing Length pEnalty Right, Incentivizing More Intelligence
  per Token via Reinforcement Learning}.
\newblock \emph{arXiv preprint arXiv:2510.15110}, 2025.

\bibitem[Luo et~al.(2025{\natexlab{a}})]{luo2025}
H.~Luo et~al.
\newblock {O1-Pruner: Length-Harmonizing Fine-Tuning for O1-Like Reasoning
  Pruning}.
\newblock \emph{arXiv preprint arXiv:2501.12570}, 2025{\natexlab{a}}.

\bibitem[Luo et~al.(2025{\natexlab{b}})]{deepscaler2025}
Michael Luo et~al.
\newblock {DeepScaleR: Surpassing O1-Preview with a 1.5B Model by Scaling RL}.
\newblock Notion Blog, 2025{\natexlab{b}}.

\bibitem[{Mathematical Association of America}(2023)]{amc}
{Mathematical Association of America}.
\newblock {AMC: American Mathematics Competitions}.
\newblock \url{https://maa.org/student-programs/amc/}, 2023.
\newblock Problem sets accessed via public community releases.

\bibitem[{Mathematical Association of America}(2025)]{aime}
{Mathematical Association of America}.
\newblock {AIME: American Invitational Mathematics Examination (2024, 2025)}.
\newblock \url{https://maa.org/student-programs/amc/}, 2025.
\newblock Problem sets accessed via public community releases.

\bibitem[Muennighoff et~al.(2025)]{muennighoff2025}
N.~Muennighoff et~al.
\newblock {s1: Simple Test-Time Scaling}.
\newblock In \emph{Proceedings of the 2025 Conference on Empirical Methods in
  Natural Language Processing (EMNLP)}, pp.\  20275--20321, 2025.
\newblock arXiv:2501.19393.

\bibitem[Rimsky et~al.(2024)Rimsky, Gabrieli, Schulz, Tong, Hubinger, and
  Turner]{rimsky2024}
N.~Rimsky, N.~Gabrieli, J.~Schulz, M.~Tong, E.~Hubinger, and A.~M. Turner.
\newblock {Steering Llama 2 via Contrastive Activation Addition}.
\newblock In \emph{Annual Meeting of the Association for Computational
  Linguistics (ACL)}, 2024.
\newblock arXiv:2312.06681.

\bibitem[Sui et~al.(2025)]{sui2025}
Y.~Sui et~al.
\newblock {Stop Overthinking: A Survey on Efficient Reasoning for Large
  Language Models}.
\newblock \emph{Transactions on Machine Learning Research (TMLR)}, 2025.
\newblock arXiv:2503.16419.

\bibitem[Turner et~al.(2023)]{turner2023}
A.~M. Turner et~al.
\newblock {Steering Language Models with Activation Engineering}.
\newblock \emph{arXiv preprint arXiv:2308.10248}, 2023.

\bibitem[Wei et~al.(2022)]{wei2022}
J.~Wei et~al.
\newblock {Chain-of-Thought Prompting Elicits Reasoning in Large Language
  Models}.
\newblock In \emph{Advances in Neural Information Processing Systems
  (NeurIPS)}, 2022.
\newblock arXiv:2201.11903.

\bibitem[Wortsman et~al.(2022)]{wortsman2022}
M.~Wortsman et~al.
\newblock {Model Soups: Averaging Weights of Multiple Fine-Tuned Models
  Improves Accuracy without Increasing Inference Time}.
\newblock In \emph{International Conference on Machine Learning (ICML)}, 2022.
\newblock arXiv:2203.05482.

\bibitem[Xu et~al.(2025)Xu, Xie, Zhao, and He]{xu2025}
S.~Xu, W.~Xie, L.~Zhao, and P.~He.
\newblock {Chain of Draft: Thinking Faster by Writing Less}.
\newblock \emph{arXiv preprint arXiv:2502.18600}, 2025.

\bibitem[Yadav et~al.(2023)]{yadav2023}
P.~Yadav et~al.
\newblock {TIES-Merging: Resolving Interference When Merging Models}.
\newblock In \emph{Advances in Neural Information Processing Systems
  (NeurIPS)}, 2023.
\newblock arXiv:2306.01708.

\bibitem[Yang et~al.(2025)]{yang2025}
C.~Yang et~al.
\newblock {Dynamic Early Exit in Reasoning Models}.
\newblock \emph{arXiv preprint arXiv:2504.15895}, 2025.

\bibitem[Zhang et~al.(2025)]{zhang2025}
A.~Zhang et~al.
\newblock {Reasoning Models Know When They're Right: Probing Hidden States for
  Self-Verification}.
\newblock In \emph{Conference on Language Modeling (COLM)}, 2025.
\newblock arXiv:2504.05419.

\bibitem[Zou et~al.(2023)]{zou2023}
A.~Zou et~al.
\newblock {Representation Engineering: A Top-Down Approach to AI Transparency}.
\newblock \emph{arXiv preprint arXiv:2310.01405}, 2023.

\end{thebibliography}
\bibliographystyle{colm2026_conference}

\appendix

\section{Why outcome-RL and imitation fail}\label{app:negatives}

This appendix gives the full methodology behind the negative results summarized in Section~\ref{sec:intro}: two offline signal preflights, a decisive training run and a confirmation that no outcome reward suffices, and five supervised and preference objectives that teach the halt by demonstration. Throughout, the training algorithm for the reinforcement-learning arms is GRPO, which for each problem samples a group of eight attempts, computes an advantage for each attempt as its score minus the group mean, and pushes the model toward the above-average attempts. One property of this scheme drives much of what follows: if all eight attempts for a problem receive the same score, the advantage spread is zero and the problem contributes no gradient. We therefore begin offline, by scoring cached attempts under candidate reward definitions and counting, for each problem, whether it is dead, meaning that all eight attempts score alike, or useful, meaning that it has at least two correct attempts of different lengths so that the reward can prefer the shorter one. This lets us judge whether a reward has any trainable signal before spending compute on a run.

\subsection{A shortness reward needs a per-problem budget}\label{app:shortness}
The first question is whether a shortness reward gives GRPO a non-flat signal at all. We answer it offline, on 3,000 cached problems with eight attempts each and no training. The pool is easy-skewed, with 56.8\% of problems solved on all eight attempts and 17.4\% solved on none, and the correct-answer length is coupled to difficulty, rising from about 2,342 tokens on easy problems to about 7,365 on hard ones. Against this pool, a single global length budget, one token limit applied to every problem, leaves roughly 52\% of problems dead, because one limit is hopelessly wrong for a set that mixes 2,000-token and 7,000-token problems. A per-problem budget, defined as the 30th percentile of each problem's own correct-answer lengths, flips the same data to roughly 1\% dead and 48\% useful, with a correlation between advantage and length among correct rollouts of about $-$0.60, so shorter correct answers score higher. A shortness reward therefore carries a trainable signal only under a per-problem budget. Two caveats bound this: a signal offline does not guarantee that training will follow it, and per-problem length budgets are already published, so we treat this as a known-good baseline.

\subsection{Saturation-point budgets look dead on easy problems and real on hard ones}\label{app:saturation}
The premise is not to budget by generic length but by the saturation point, the moment at which the model appears internally confident, so it is asked to stop where it already knows. Using the same offline machinery, we check three properties of a saturation-derived budget: reachability, whether a correct answer that short exists; usefulness, whether the reward is dead; and orthogonality to plain length, whether it adds anything new. On the mostly-easy DeepScaleR pool the signal looks dead and unreachable: a correct answer as short as the saturation point exists for only about 10.5\% of problems against 100\% for a length budget, roughly 48\% of groups are dead, the saturation budget is about 13\% of the length budget, the advantage correlation is near zero and wrong-signed at about $+$0.09 against length's $-$0.42, and saturation is roughly uncorrelated with the length actually needed, between $-$0.03 and $+$0.14. On the harder AMC set the picture reverses: the model saturates only after about 1,570 tokens of genuine reasoning, with the budget at about half the correct length. Saturation therefore looks dead on easy problems and promising on hard ones.

\subsection{Outcome RL only trims the tail}\label{app:outcomerl}
Two refinements make the saturation budget trustworthy before the decisive run. First, the definition of knowing must be strict: a match on the first token of the gold answer fires before the model has done any work, so it is an artifact, and the correct replacement is forced-answer correctness, which on AMC is right 86\% of the time at the saturation point against only 16\% with zero reasoning, so reasoning adds about 70 points and AMC saturation is about 94\% earned rather than a format prior. Second, one should not budget directly at the saturation point, whose gradient is weak at a correlation of about $-$0.09, but should anneal a reachable correct-length budget toward it. As a metric note, whole-answer probes read zero on AMC purely because of a LaTeX string-matching artifact, so a forced-answer check with \texttt{math\_verify} is the correct cross-dataset metric.

With those in place, we train a per-problem budget that anneals from each problem's reachable start, the median of its own correct lengths, down toward its forced-answer point, with a reward floor so that a correct but long answer never scores below a wrong one. Seven arms run in parallel: the full DeepScaleR set, a goldilocks subset, a hard subset, AMC, versions with and without a forced-answer floor, and a naive global length-penalty control. The clean results at a 16k cap are in Table~\ref{tab:outcomerl}.

\begin{table}[t]
\centering
\begin{tabular}{lrrr}
\toprule
arm (AMC, 16k, $n=8$) & accuracy & mean think-tok & $\Delta$think \\
\midrule
base & 0.813 & 6280 & n/a \\
naive length penalty (control) & 0.803 & 5911 & $-$5.9\% \\
per-problem budget (best) & 0.830 & 5996 & $-$4.5\% \\
per-problem budget (full data) & 0.792 & 6264 & $-$0.3\% \\
\bottomrule
\end{tabular}
\caption{\textbf{Outcome-RL on AMC.} Per-problem saturation budgets against a naive global length penalty (16k cap, $n=8$). The global penalty beats every per-problem budget, and compression is small.}
\label{tab:outcomerl}
\end{table}

Three findings follow. First, accuracy is preserved everywhere, between 0.79 and 0.83, but compression is small, between 0 and 6\%, and the naive global penalty beats every per-problem budget, so the elaborate saturation curriculum bought nothing over the simplest possible penalty. Second, the internal saturation point does not move, staying near 1,650 tokens across every arm including the base model, and what compression the length penalty achieves is confined to the tail. Over all rollouts its median emitted length is 7.7\% \emph{longer} than base, 4,414 tokens against 4,100. The saving sits instead on the rollouts that saturate and then talk themselves into a wrong answer, 6.6\% of the set: there the tokens emitted after the internal crossing fall from 6,369 to 2,289, a 64\% reduction ($n=44$ and $n=45$). Measured across all saturated rollouts, the same post-crossing waste falls 18.9\%. Either way the penalty removes the longest ramblers without ever teaching the median rollout to stop earlier. Third, there is a striking dissociation between training and deployment. During training the budget arms look healthy, with an advantage-length correlation between $-$0.9 and $-$0.99, near-zero dead groups, and the budget annealing down by 30 to 60\%, and yet the deployed model reasons for as long as the control. A boundary condition also holds: a problem the model can barely solve cannot be compressed. The interpretation is that outcome-level reinforcement learning, whether a length penalty or an elaborate per-problem saturation budget, only trims the tail, because the reward is a single number for a 5,000-token trace and the decision of where to stop is smeared across thousands of tokens, which is poor credit assignment.

\subsection{Even a perfectly conditioned outcome reward only trims the tail}\label{app:conditioned}
To rule out that the failure was merely mechanical, we removed every remaining excuse at once: an additive constant-slope reward whose training signal is provably independent of how far the budget has annealed, a reachable target with the forced-answer point set to the 20th percentile of each problem's own correct lengths, and a floor so that a correct answer always outscores a wrong one. The result is unchanged. This confirms the decisive run, which already showed that the internal point is fixed and any outcome reward can only trim the tail; it is the clean statement that even a perfectly conditioned outcome reward does not localize the stop, which is what motivates supervising the stop directly.

\subsection{Imitation and preference optimization fail too}\label{app:imitation}
If reshaping an outcome reward cannot localize the stop decision, the alternative is to place the supervision directly at the stop. We take the model's own correct traces, cut each at its forced-answer point plus a small margin, staple the answer on, and train on the result, so that every example demonstrates that the model knew the answer here and therefore stopped here. Anchor problems, which have no early answer point, are kept at full length so that the model learns when not to stop. The resulting dataset has 1,300 rows, 900 halt examples and 400 anchors. One reproducibility caution: the DeepSeek-R1 chat template silently strips the \texttt{<think>} and \texttt{</think>} tags from assistant messages, which trains the model to answer with no reasoning and collapses accuracy, so the training text must be composed manually. Five families of objectives follow (Table~\ref{tab:families}).

\begin{table}[t]
\centering
\begin{tabular}{@{}>{\raggedright\arraybackslash}p{3.3cm}>{\raggedright\arraybackslash}p{4.8cm}>{\raggedright\arraybackslash}p{4.6cm}@{}}
\toprule
method & where the ``stop'' signal lands & outcome \\
\midrule
\textbf{outcome RL} & one scalar over ${\sim}5{,}000$ tokens & tail-trims only \\
\addlinespace
\textbf{plain SFT} \emph{(imitate short traces)} & 0.34\% of the token loss & no change in length \\
\addlinespace
\textbf{SimPO} \emph{(length-normalized preference)} & canceled by length-normalization & null \\
\addlinespace
\textbf{vanilla DPO} \emph{(prefer short)} & a ranking of two fixed sequences & held accuracy, no compression \\
\addlinespace
\textbf{boundary-weighted SFT} \emph{(up-weight stop tokens)} & on the stop tokens & moves length, trades accuracy ${\sim}1{:}1$ \\
\bottomrule
\end{tabular}
\caption{\textbf{Where the stop signal lands for each objective.} Every family either puts no weight on the stopping decision or, when forced to, installs a single global cut that trades accuracy.}
\label{tab:families}
\end{table}

Each mechanism is a clean lesson. Plain supervised fine-tuning does nothing, because the stop-and-commit tokens are 0.34\% of a loss averaged over all tokens, so they are invisible to a loss dominated by the reproduction of reasoning the model already writes. Evaluated on its own training problems, the fine-tuned adapter actually gets longer, at 7,174 tokens against 6,505, so the objective failed to learn in-distribution; transfer was never the bottleneck. SimPO is null because it scores by average per-token probability, and our short and long pairs share the same fluent reasoning per token, so length-normalization cancels the length signal exactly; the practical lesson is that one should never use a length-normalized preference method for a length objective. Vanilla DPO shows a near-perfect training curve, with a reward margin reaching 5.2 and preference accuracy reaching 100\%, that does not transfer at all, an effect known as likelihood displacement: because the chosen short sequence is a prefix of the rejected long one, pushing the long sequence down drags the short one down as well, and probability flows to other long continuations, so the model learns to rank two fixed essays rather than to write a short one. Boundary-weighted SFT, which up-weights the last roughly 128 stop tokens, is the only objective that moves length, but it has no knee, so every bit of compression costs accuracy at roughly one point per point: a weight of 2 gives a 13\% length reduction at a 4-point accuracy cost, while a weight of 10 gives a 49\% reduction at a 24-point cost, and the damage concentrates on the hardest problems. It installs a global rule to stop around some length rather than a per-problem sense of whether the answer is known yet.

\subsection{The unifying reason and the in-distribution gate}\label{app:unifying}
Across these objectives the overthinking gap is large and reachable, but it is not installable by shaping an outcome reward, nor by imitation, nor by preference optimization. The common reason is that every objective either puts no weight on the stopping decision, or, when forced to, installs a single global cut that trades accuracy, and none of them supplies a per-problem sensor that conditions stopping on the model's own knowing-state. This negative also yields the evaluation gate we rely on in the main text. Since no method produced a shorter output at held accuracy even on its own training problems, the right first question for any new method is not whether it generalizes but whether it can produce a shorter output at accuracy on a small, deliberately overfit training set at all. A method that cannot clear this in-distribution gate has no purchase, and a method that can has isolated a real learnable signal, at which point the question becomes transfer.

\section{Implementation and reproducibility}\label{app:impl}

We record the details needed to reproduce the results. All training and evaluation ran on Modal with H200 GPUs. The headline adapter is a single-GPU supervised fit of an attention-only LoRA on layers 0 to 18, and evaluation uses a sampling decoder at a 16k-token cap, with no reinforcement-learning infrastructure used for the headline result. Accuracy is measured with \texttt{math\_verify}, which is format-robust, and we report comparator agreement and lean on the grader-independent length signal wherever a legacy comparator disagrees. We use the same 16k cap for every arm, because a cap that truncates reasoning corrupts both accuracy and the length metric, and we average held-out numbers over 4 to 16 samples with bootstrap-over-problems 95\% confidence intervals, judging every arm on its deployed behavior. One caution for anyone replicating this work: the DeepSeek-R1 chat template silently strips the \texttt{<think>} and \texttt{</think>} tags from assistant messages, so the training text must be composed manually (Appendix~\ref{app:negatives}). Finally, we reimplemented DEER end to end rather than relying on its published numbers, so that the comparison shares our grader and benchmarks.

\section{How the halt vector is constructed}\label{app:axis}

Section~\ref{sec:background} summarizes the construction; we give it in full here, because the
instrument that selects the contrast is not the instrument the rest of the paper cuts on.

\paragraph{The contrast.} We take rollouts from the base model on the offline pool and keep only
those the model answers correctly, since the lock-on point is undefined for a rollout that never
reaches the right answer. For each kept rollout we split the thinking into chunks and score every
chunk prefix with the same forced-answer machinery used for \bend, recording the model's
probability on the first token of the gold answer. The lock-on chunk is the first chunk at which
that probability reaches 0.9 \emph{and stays at or above 0.9 for every later chunk}; requiring it
to stay high rejects transient spikes. Rollouts that never lock, and rollouts already locked at the
first chunk (which would leave an empty ``before'' set), are dropped. The remaining rollouts each
contribute one item to the contrast set $C$ of Equation~\ref{eq:diffmeans}, with $T_{\text{post}}$
the think-token positions at or after the lock-on point and $T_{\text{pre}}$ those before it. The
contrast is therefore \emph{within} a rollout, so the problem, its difficulty, and the trace length
are held fixed across the two sides by construction.

\paragraph{Census.} The probe pool is 664 rollouts over 83 problems. Of these, 480 are answered
correctly and so admit a lock-on point; 457 of those yield a usable contrast, with 11 dropped
because the probability never locks on and 12 because it is already locked at the first chunk,
which would leave an empty ``before'' set. The direction is built from those 457. One property of
the trajectories is worth stating, since it bounds how the landmark should be read: in 131 of the
480 correct rollouts (27\%) the probability crosses 0.9 and later falls back before finally
settling. The stay-above rule places the landmark at the final settling point in those cases, so
lock-on marks where the probability last becomes stable, not the first moment the model is
transiently confident.

\paragraph{What the AUROC measures, and on which split.} The direction is accumulated over a
construction split and scored on a held-out split. The score is computed at the level of
\emph{token positions}: each held-out post-lock-on position is labelled 1, each pre-lock-on
position 0, and the score is that position's projection onto the unit direction. It measures how
well the direction separates late reasoning positions from early ones. It is not a rollout-level
discrimination of correct from incorrect answers and should not be read as one.

Two limits on that number. First, the split key is applied per \emph{rollout}, not per problem, and
with eight samples per problem it places every one of the 83 problems on both sides. The held-out
score therefore measures generalization across fresh samples of problems the direction has seen,
not across unseen problems, and a problem-disjoint split would be expected to score lower. Second,
we report the value at layer 18, which is where the direction is used for everything else in this
paper. The maximum over all 28 layers is higher, 0.80 at layer 6, but that is an argmax selected on
the same held-out set and is not comparable to a value read off at a layer fixed in advance. Layer
6 reading best while layer 18 steers best (Section~\ref{sec:causal}) is not a contradiction: how
legible a feature is to a linear read-out and how well an intervention on it propagates through the
remaining layers are different questions.

\paragraph{The value axis, measured the same way.} The value axis's number is the same kind of
quantity, and it is worth stating in full because the naive comparison runs the wrong way. Its
across-layer maximum is 0.783 at layer 25, but unlike the halt vector the selection barely moves
it: at layer 18 it reads 0.780, the same 0.78 to the precision we print. Its split is also cleaner.
It is held out by conversation, and the unit that actually repeats across conversations, the
criterion, is disjoint: of 50 distinct criteria, 34 are construction-only and 16 held-out-only,
with none on both sides. Its held-out sample is 94 conversations against the halt vector's 457
rollouts.

So at layer 18, on each direction's own held-out split, the value axis reads 0.780 and the halt
vector 0.758. The control separates slightly \emph{better}, on the stricter split. We report this
plainly because it is easy to assume the opposite from the ordering of the two numbers in
Section~\ref{sec:background}. It does not weaken the paper's claim about the two directions, which
is a claim about steering rather than read-out: the halt vector moves generation length and the
value axis does not (Sections~\ref{sec:causal} and~\ref{sec:dirspecific}), and neither of those
results depends on an AUROC. What it does rule out is any reading in which the halt vector is
selected for being the more legible feature. It is not.

One shared caveat: on both directions the stored item count describes the input pool, not the
scored sample. The AUROCs above are over 457 held-out lock-on rollouts and 94 held-out
conversations respectively, not over the 664 and 300 rows the artifacts record.

\paragraph{A between-item variant that we rejected.} We also built the direction the obvious other
way, as a between-item contrast of answer tokens from correct rollouts against answer tokens from
incorrect ones. It scores far higher, and that is precisely why we do not use it. Its held-out
AUROC is 0.955 \emph{at layer 0}, and the curve is flat near 0.95 across all 28 layers
(Table~\ref{tab:auroc}). Layer 0 is the embedding layer, before the network has computed anything,
so a probe that already separates there is separating on properties of the problem text, and a
curve that stays flat with depth is one to which depth contributes nothing. The lock-on contrast
behaves in the opposite way: 0.573 at layer 0, rising through the middle of the network. That
direction is not available at the input and has to be built, which is what we want from a feature
that is supposed to develop during reasoning.

\begin{table}[t]
\centering
{\footnotesize
\setlength{\tabcolsep}{3.2pt}
\begin{tabular}{l*{14}{r}}
\toprule
layer & 0 & 1 & 2 & 3 & 4 & 5 & 6 & 7 & 8 & 9 & 10 & 11 & 12 & 13 \\
\midrule
lock-on (ours) & .573 & .658 & .685 & .788 & .795 & .785 & \textbf{.804} & .795 & .795 & .803 & .792 & .773 & .764 & .768 \\
between-item & .955 & .961 & .960 & .945 & .943 & .945 & .946 & .942 & .942 & .943 & .942 & .945 & .946 & .947 \\
\midrule
layer & 14 & 15 & 16 & 17 & \textbf{18} & 19 & 20 & 21 & 22 & 23 & 24 & 25 & 26 & 27 \\
\midrule
lock-on (ours) & .761 & .768 & .753 & .748 & \textbf{.758} & .776 & .793 & .782 & .787 & .799 & .796 & .793 & .766 & .554 \\
between-item & .949 & .950 & .950 & .950 & .951 & .952 & .949 & .953 & .953 & .958 & \textbf{.965} & .961 & .954 & .955 \\
\bottomrule
\end{tabular}}
\caption{\textbf{Held-out AUROC by layer for both math contrasts.} The lock-on contrast we use
starts near chance at the embedding layer and has to be built with depth; layer 18, used throughout
the paper, reads 0.758, and the across-layer maximum of 0.804 at layer 6 is an argmax selected on
this same held-out set. The rejected between-item contrast is already at 0.955 at layer 0 and flat
thereafter, the signature of a surface confound rather than a computed feature.}
\label{tab:auroc}
\end{table}

\paragraph{Relation to the forced-answer point.} The first-token probability used here is the
signal Appendix~\ref{app:outcomerl} rejects as a measure of knowing, on the grounds that it can
fire on the answer-format prior before any reasoning has happened. That objection is about using it
as a \emph{target}, and we do not: \bend, every training target, the budget references, the
goldilocks and anchor classification, and the reconstruction masks are all forced-answer
quantities. The first-token probability appears only as a cheap positional landmark for placing a
difference-of-means contrast, and whether the resulting direction is any good is settled by the
causal steering test of Section~\ref{sec:causal}, not by how the contrast was placed.

\section{Additional tables}\label{app:tables}

This appendix collects the detailed tables behind the results summarized in the main text. All rows use a 16k-token cap. Accuracy is \texttt{math\_verify}, tok\_think is the mean think-token count, and closed is the fraction that emitted \texttt{</think>}. Steering strength is written $\alpha$ throughout.

\subsection{The reconstruction objective: schedule, strength, and capacity}\label{app:recongrid}
The headline adapter is selected from a grid over the post-\bend schedule (hold, the gated variant, and tent), the steering strength, and the adapter capacity (attention-only against attention-plus-MLP), all on the 24 compressible problems against a base of 0.812 accuracy, 6,516 think-tokens, and 0.95 closure. Section~\ref{sec:internalize} states the three ablation conclusions; the grid behind them is in Table~\ref{tab:recongrid}.

\begin{table}[t]
\centering
\begin{tabular}{lrrrr}
\toprule
arm & acc & tok\_think & $\Delta$think & closed \\
\midrule
base & 0.812 & 6516 & n/a & 0.95 \\
hold $\alpha=25$, attn (headline) & 0.792 & 4911 & $-$25\% & 0.99 \\
hold $\alpha=10$, attn & 0.833 & 5818 & $-$11\% & 0.96 \\
hold $\alpha=10$, attn+MLP & 0.698 & 5235 & $-$20\% & 0.99 \\
hold $\alpha=25$, attn+MLP & 0.740 & 5943 & $-$9\% & 0.95 \\
gate $\alpha=10$, attn & 0.823 & 5870 & $-$10\% & 0.93 \\
tent $\alpha=10$, attn & 0.781 & 6470 & $-$1\% & 0.94 \\
\bottomrule
\end{tabular}
\caption{\textbf{Reconstruction grid: schedule, strength, capacity.} Selection grid on the 24 compressible problems (base: 0.812 acc, 6{,}516 think-tokens, 0.95 closed). Hold and gate install the halt; attention-only beats attention-plus-MLP.}
\label{tab:recongrid}
\end{table}

The hold and gate schedules install the halt; tent barely moves length. Attention-only beats attention-plus-MLP at both strengths, because the extra capacity lets the writer distort more off-axis dimensions.

The scalar-projection target fails in the opposite direction: the harder the writer is pushed to raise the scalar projection, the longer and worse the model becomes. Table~\ref{tab:scalarproj} reports the achieved post-\bend projection against length on the same 24 problems, where the target margin is the size of the scalar target.

\begin{table}[t]
\centering
\begin{tabular}{lrrrrr}
\toprule
scalar-projection arm & proj\_post & acc & tok\_think & $\Delta$think & closed \\
\midrule
base & n/a & 0.812 & 6516 & n/a & 0.95 \\
attn+MLP, target margin 50 & $+$0.336 & 0.573 & 9331 & $+$43\% & 0.61 \\
raw+MLP, target margin 30 & $+$0.238 & 0.573 & 8984 & $+$38\% & 0.67 \\
attn+MLP, target margin 15 & $+$0.119 & 0.781 & 7648 & $+$17\% & 0.85 \\
raw, target margin 15 & $+$0.077 & 0.688 & 7453 & $+$14\% & 0.79 \\
\bottomrule
\end{tabular}
\caption{\textbf{The scalar-projection target fails.} Achieved post-\bend projection against length on the 24 problems. Pushing the projection harder makes the model longer and less accurate, the off-axis-corruption signature.}
\label{tab:scalarproj}
\end{table}

The two arms that actually raise the projection past $+$0.13 are the highest-capacity ones, and they are also the longest and least accurate.

\subsection{Off-axis drift by arm}\label{app:drift}
Section~\ref{sec:internalize} claims that a scalar target lets the writer disturb the off-axis
dimensions while reconstruction does not. Table~\ref{tab:drift} measures it. At layer 18 each
think-token activation is decomposed against the unit halt vector into
$h_{\parallel} = (h \cdot \hat u)\hat u$ and $h_{\perp} = h - h_{\parallel}$, and compared to the
base activation at the same position. \emph{on-axis} is the median achieved move in units of
$\alpha/100$; \emph{drift} is the median $\lVert h_{\perp} - h_{\perp,\text{base}}\rVert$ per unit
steer magnitude; \emph{\%out} is the fraction of think tokens whose drift exceeds the base model's
own p99 off-axis band; \emph{KL} is the median next-token KL of the adapter against the base on the
same tokens, which is the downstream-reader sensitivity the geometry alone does not establish. All
values come from one teacher-forced pass with the adapter enabled and one with it disabled, so no
generation is involved.

\begin{table}[t]
\centering
\small
\begin{tabular}{llrrrrr}
\toprule
arm & family & on-axis & drift & drift/on-axis & \%out & KL p50 \\
\midrule
null & control & $+$0.0000 & 0.000 & n/a & 0.0\% & 0.0000 \\
\addlinespace
reconmlp & reconstruction & $+$0.1292 & 0.272 & 2.11 & 0.0\% & 0.0004 \\
champion & reconstruction & $+$0.0828 & 0.303 & 3.66 & 0.0\% & 0.0002 \\
\addlinespace
m15 & scalar, target & $+$0.1474 & 0.769 & 5.22 & 0.2\% & 0.0005 \\
m50lam4 & scalar, $\lambda$ & $+$0.1491 & 0.848 & 5.69 & 1.1\% & 0.0006 \\
m50lam2 & scalar, $\lambda$ & $+$0.1257 & 0.718 & 5.71 & 0.3\% & 0.0005 \\
m30 & scalar, target & $+$0.0996 & 0.569 & 5.71 & 0.2\% & 0.0005 \\
m50lam8 & scalar, $\lambda$ & $+$0.1453 & 0.840 & 5.78 & 1.1\% & 0.0006 \\
rawmlpm30 & scalar, capacity & $+$0.1541 & 0.949 & 6.16 & 0.4\% & 0.0011 \\
m50 & scalar, target & $+$0.0754 & 0.465 & 6.17 & 0.2\% & 0.0005 \\
mlpm50 & scalar, capacity & $+$0.1882 & 1.188 & 6.31 & 4.0\% & 0.0015 \\
\bottomrule
\end{tabular}
\caption{\textbf{Off-axis drift by arm.} Reconstruction and scalar-projection arms separate with no
overlap on all four measures: raw drift (0.303 against 0.465), the ratio to achieved on-axis
movement (3.66 against 5.22), \%out, and reader KL (0.0004 against 0.0005).}
\label{tab:drift}
\end{table}

Three readings, and one non-reading. First, the objection that reconstruction drifts less merely
because it moves less is answered by the ratio column: normalizing by achieved on-axis movement,
the separation widens rather than closes. Second, the separation is fourfold and has no overlap on
any of the four measures, including reader KL. Third, drift tracks behavioural damage across arms:
the rank correlation between drift and training-set accuracy is $\rho = -0.799$ (midrank Spearman,
exact two-sided permutation $p = 0.0079$, and $p = 0.032$ under a fourfold Bonferroni correction).

The non-reading is that none of this shows the scalar arms leave the reader's natural off-axis
range. Normalized against the base model's own off-axis dispersion, every arm including the worst
sits below the base's mean (largest $z = -1.17$), and on \%out no arm other than \texttt{mlpm50}
exceeds 1.1\% against the base model's own 1\% excursion rate, so only \texttt{mlpm50} at 4.0\% is
clearly above it. The result is comparative, not categorical. We also measured a random-direction arm and omit it
here: it was scored against the halt axis rather than against its own, so its drift is its entire
edit counted as drift and is not comparable to the rows above.

\subsection{Adding an off-axis penalty to the scalar objective: an uninformative null}\label{app:perp}
Section~\ref{sec:internalize} shows that a full-vector target with off-axis pinned works and a
scalar target without it does not. That leaves a crossed design with one cell unfilled: a scalar
target \emph{with} off-axis constrained. We ran it. The arm is the scalar arm \texttt{m15}
unchanged, plus a term penalizing $\lVert h_{\perp} - h_{\perp,\text{base}}\rVert^2$ over the think
tokens, normalized by the same steer magnitude as the reconstruction loss so its weight
$\lambda_{\perp}$ is on a comparable scale. Everything else is the champion recipe: layer 18, seed
17, 12 epochs, attention-only LoRA on layers 0 to 18, the same 24 problems. We fixed in advance
that the arm had to bring drift into the reconstruction band, at or below 3.66 on drift per unit
achieved on-axis movement \emph{and} at or below the scalar family's 0.465 on raw drift, before we
would evaluate its held-out behaviour at all.

\begin{table}[t]
\centering
\small
\begin{tabular}{lrrrrr}
\toprule
arm & on-axis & drift & drift/on-axis & \%out & KL p50 \\
\midrule
champion (reference) & $+$0.0828 & 0.303 & 3.66 & 0.0\% & 0.0002 \\
\texttt{m15}, no penalty & $+$0.1474 & 0.769 & 5.22 & 0.2\% & 0.0005 \\
\addlinespace
\texttt{m15} $+$ penalty, $\lambda_{\perp}=1$ & $+$0.0120 & 0.053 & 4.43 & 0.0\% & 0.0000 \\
\texttt{m15} $+$ penalty, $\lambda_{\perp}=4$ & $+$0.0046 & 0.044 & 9.46 & 0.0\% & 0.0000 \\
\texttt{m15} $+$ penalty, $\lambda_{\perp}=16$ & $+$0.0008 & 0.041 & 50.91 & 0.0\% & 0.0000 \\
\bottomrule
\end{tabular}
\caption{\textbf{Scalar target with an off-axis preservation penalty.} Raw drift falls an order of
magnitude, but achieved on-axis movement falls with it, so the ratio does not enter the
reconstruction band and no arm was carried through to a held-out evaluation.}
\label{tab:perp}
\end{table}

No weight passed. Raw drift does fall, and dramatically, from 0.769 to 0.041, far below the 0.465
floor of the whole scalar family. But achieved on-axis movement falls with it, from $+$0.1474 to
$+$0.0008, so the ratio moves the wrong way and never enters the reconstruction band. At
$\lambda_{\perp}=16$ the arm achieves half a percent of the unpenalized arm's on-axis movement,
with \%out and reader KL indistinguishable from making no edit at all. The penalty did not
selectively remove off-axis drift; it suppressed the edit.

We therefore report this as \textbf{uninformative} rather than as evidence for either side. It does
not show that an off-axis penalty cannot rescue scalar training, because the arms never reached a
state in which that question could be asked; what it weakly suggests is that this penalty, at these
weights, trades on-axis achievement against off-axis cleanliness at close to one for one. The
crossed design still has an empty cell: a scalar target with off-axis constrained \emph{and}
on-axis preserved was not realized. One observation we record without drawing anything from it: the
ratio is not monotone in $\lambda_{\perp}$, improving on the unpenalized arm at $\lambda_{\perp}=1$
before degrading. We deliberately did not extend the sweep downward after seeing this. Choosing a
new weight in the direction a failed pre-registration points is how such a test stops being one,
and a smaller weight would need its own criterion fixed in advance, stating what a pass would
license given that the visible mechanism here is less learning rather than cleaner learning.

\paragraph{In-distribution method comparison and the anchor split.} Table~\ref{tab:indist} gives the full in-distribution comparison behind Section~\ref{sec:fit}, and Table~\ref{tab:anchorsplit} the full compressible-versus-anchor strength sweep behind the adaptivity claim there.

\begin{table}[t]
\centering
\begin{tabular}{lrrrrr}
\toprule
method & acc & $\Delta$acc (pp) & tok\_think & $\Delta$think & closed \\
\midrule
base & 0.812 & n/a & 6516 & n/a & 0.95 \\
imitation (halt-SFT) & 0.708 & $-$10.4 & 7500 & $+$15\% & 0.82 \\
steering-distillation (uniform) & 0.740 & $-$7.2 & 6057 & $-$7\% & 0.85 \\
steering-distillation (gated) & 0.750 & $-$6.2 & 5179 & $-$21\% & 0.93 \\
\textbf{reconstruction (ours)} & 0.792 & $-$2.0 & 4911 & $-$25\% & 0.99 \\
\bottomrule
\end{tabular}
\caption{\textbf{In-distribution comparison on the 24 training problems.} All rows use a 16k-token cap and $n=4$ samples. \emph{acc}: \texttt{math\_verify} accuracy; \emph{tok\_think}: mean think-token count; \emph{closed}: fraction emitting \texttt{</think>}; pp: percentage points. Only reconstruction compresses at held accuracy.}
\label{tab:indist}
\end{table}

\begin{table}[t]
\centering
{\small
\begin{tabular}{rrrrrrr}
\toprule
$\alpha$ & comp acc & comp think & comp $\Delta$think & anchor acc & anchor think & anchor $\Delta$think \\
\midrule
10 & 0.833 & 5818 & $-$11\% & 0.719 & 4912 & $+$2\% \\
25 & 0.792 & 4911 & $-$25\% & 0.750 & 4174 & $-$13\% \\
40 & 0.781 & 4788 & $-$27\% & 0.812 & 3909 & $-$19\% \\
50 & 0.740 & 4511 & $-$31\% & 0.750 & 3778 & $-$21\% \\
\bottomrule
\end{tabular}}
\caption{\textbf{The steering-strength dial: compressible versus anchor.} Full strength sweep on both sets (compressible base 0.812 acc / 6{,}516 think; anchor base 0.719 / 4{,}812). At $\alpha=10$ the anchors are essentially untouched ($+$2\%) while the compressible set is cut 11\%; anchor accuracy never falls below base, so the higher-$\alpha$ anchor cuts remove probe-missed slack.}
\label{tab:anchorsplit}
\end{table}

\subsection{The truncation family}\label{app:truncation}
Truncating each training trace at \bend removes the erosion that afflicts long single runs but spends accuracy to buy length. Hard truncation, which deletes the post-\bend tokens, is in Table~\ref{tab:hardtrunc} on the goldilocks holdout (base 0.845 / 6617 / 0.93) and cross-benchmark AMC (base 0.813 / 6280 / 0.89), with $\lambda$ the reconstruction weight and every arm trained on 150 problems.

\begin{table}[t]
\centering
\small
\setlength{\tabcolsep}{4pt}
\begin{tabular}{lrrrrrr}
\toprule
 & \multicolumn{3}{c}{holdout} & \multicolumn{3}{c}{AMC} \\
\cmidrule(lr){2-4}\cmidrule(lr){5-7}
adapter (150 problems) & acc & $\Delta$think & closed & acc & $\Delta$think & closed \\
\midrule
recon-full, ep4 & 0.792 & $-$19\% & 0.96 & 0.780 & $-$20\% & 0.96 \\
recon-full, ep12 (erosion) & 0.825 & $-$0.2\% & 0.93 & 0.788 & $-$9\% & 0.96 \\
hybrid $\lambda=1$, ep4 & 0.790 & $-$12.6\% & 0.94 & 0.812 & $-$17.8\% & 0.95 \\
hybrid $\lambda=1$, ep12 & 0.733 & $-$23.1\% & 0.86 & 0.773 & $-$19.9\% & 0.79 \\
hybrid $\lambda=0$, ep12 (recon off) & 0.802 & $-$0.9\% & 0.92 & 0.773 & $+$1.9\% & 0.89 \\
\bottomrule
\end{tabular}
\caption{\textbf{Hard truncation (150 problems).} Deleting post-\bend tokens; $\lambda$ is the reconstruction weight. Truncation removes the erosion ($\lambda=1$, ep12) but spends accuracy, and $\lambda=0$ barely cuts, so reconstruction is the load-bearing ingredient.}
\label{tab:hardtrunc}
\end{table}

The $\lambda=0$ control barely cuts, so reconstruction is the load-bearing ingredient; the $\lambda=1$ ep12 arm removes the erosion, at $-$23.1\% where the full recipe fell to $-$0.2\%, but at an 11-point accuracy cost on the holdout. Soft truncation, which down-weights rather than deletes the post-\bend tokens, traces the same frontier (Table~\ref{tab:softtrunc}).

\begin{table}[t]
\centering
\small
\setlength{\tabcolsep}{4pt}
\begin{tabular}{lrrrrrr}
\toprule
 & \multicolumn{3}{c}{holdout} & \multicolumn{3}{c}{AMC} \\
\cmidrule(lr){2-4}\cmidrule(lr){5-7}
soft-truncation arm & acc & $\Delta$think & closed & acc & $\Delta$think & closed \\
\midrule
$w=0.10$, ep4 & 0.790 & $-$32.6\% & 0.98 & 0.800 & $-$31.5\% & 0.99 \\
$w=0.10$, ep12 & 0.780 & $-$12.3\% & 0.96 & 0.812 & $-$21.6\% & 0.98 \\
$w=0.25$, ep4 & 0.784 & $-$29.2\% & 0.97 & 0.791 & $-$27.8\% & 0.97 \\
$w=0.25$, ep12 & 0.752 & $-$6.3\% & 0.95 & 0.803 & $-$16.3\% & 0.98 \\
$w=0.50$, ep4 & 0.770 & $-$24.3\% & 0.97 & 0.789 & $-$26.1\% & 0.98 \\
$w=0.50$, ep12 & 0.772 & $-$3.3\% & 0.94 & 0.798 & $-$11.3\% & 0.97 \\
\bottomrule
\end{tabular}
\caption{\textbf{Soft truncation.} Post-\bend tokens kept at loss down-weight $w$. No holdout cell is both held-accuracy and materially compressed; $w$ is a softer knob on the same trade-off line as hard truncation.}
\label{tab:softtrunc}
\end{table}

On the discriminating holdout no soft cell is both held-accuracy and materially compressed; the down-weight $w$ is a softer margin knob on the same trade-off line. On the more forgiving AMC, $w=0.10$ at ep12 does hold accuracy at $-$21.6\%, but AMC held-accuracy cells were already reached by hard truncation. The headline therefore stays the 24-problem adapter.

\subsection{DEER per-threshold cells with greedy anchors}\label{app:deer}
We reimplement DEER \citep{yang2025} under our grader with greedy decoding at $n = 1$, sweeping the confidence threshold $\lambda$ over 0.90, 0.95, and 0.98 on the base model and on our adapter (champ). To compute a fair delta we add a never-exit greedy anchor per surface, because greedy decoding runs to the cap often enough on the hard and goldilocks sets to depress its own baseline. The $\Delta$ columns are against the matching greedy anchor (Table~\ref{tab:deerfull}).

\begin{table}[t]
\centering
{\small
\begin{tabular}{lrrrrrr}
\toprule
surface / arm & $\lambda$ & acc & $\Delta$acc (pp) & tok\_think & $\Delta$think & closed \\
\midrule
AMC, base greedy & n/a & 0.771 & n/a & 6796 & n/a & 0.77 \\
DEER base & 0.98 & 0.747 & $-$2.4 & 4236 & $-$37.7\% & 0.89 \\
DEER base & 0.95 & 0.687 & $-$8.4 & 4062 & $-$40.2\% & 0.89 \\
DEER base & 0.90 & 0.590 & $-$18.1 & 3215 & $-$52.7\% & 0.96 \\
\addlinespace
AMC, champ greedy & n/a & 0.687 & n/a & 6899 & n/a & 0.73 \\
DEER champ & 0.95 & 0.699 & $+$1.2 & 3513 & $-$49.1\% & 0.92 \\
\addlinespace
holdout, base greedy & n/a & 0.660 & n/a & 8352 & n/a & 0.66 \\
DEER base & 0.98 & 0.710 & $+$5.0 & 3764 & $-$54.9\% & 0.92 \\
DEER base & 0.95 & 0.710 & $+$5.0 & 3008 & $-$64.0\% & 0.96 \\
DEER base & 0.90 & 0.630 & $-$3.0 & 2404 & $-$71.2\% & 0.97 \\
\addlinespace
holdout, champ greedy & n/a & 0.670 & n/a & 7238 & n/a & 0.73 \\
DEER champ & 0.95 & 0.700 & $+$3.0 & 2483 & $-$65.7\% & 0.97 \\
\addlinespace
AIME24, base greedy & n/a & 0.467 & n/a & 11560 & n/a & 0.47 \\
DEER base & 0.98 & 0.433 & $-$3.4 & 8733 & $-$24.5\% & 0.67 \\
DEER base & 0.95 & 0.400 & $-$6.7 & 8536 & $-$26.2\% & 0.67 \\
DEER base & 0.90 & 0.367 & $-$10.0 & 7234 & $-$37.4\% & 0.77 \\
\addlinespace
AIME24, champ greedy & n/a & 0.400 & n/a & 9513 & n/a & 0.57 \\
DEER champ & 0.95 & 0.400 & 0.0 & 6409 & $-$32.6\% & 0.77 \\
\addlinespace
MATH500, base greedy & n/a & 0.876 & n/a & 3549 & n/a & 0.91 \\
DEER base & 0.98 & 0.878 & $+$0.2 & 1947 & $-$45.1\% & 0.97 \\
DEER base & 0.95 & 0.836 & $-$4.0 & 1629 & $-$54.1\% & 0.98 \\
DEER base & 0.90 & 0.770 & $-$10.6 & 1333 & $-$62.4\% & 0.99 \\
\addlinespace
MATH500, champ greedy & n/a & 0.886 & n/a & 2841 & n/a & 0.94 \\
DEER champ & 0.95 & 0.828 & $-$5.8 & 1415 & $-$50.2\% & 0.98 \\
\bottomrule
\end{tabular}}
\caption{\textbf{DEER per-threshold cells with greedy anchors.} Thresholds $\lambda\in\{0.90,0.95,0.98\}$ on the base model and our adapter (champ), greedy $n=1$; $\Delta$ is against the matching never-exit greedy anchor. The $\lambda=0.95$ base rows are the operating points in Table~\ref{tab:frontier}.}
\label{tab:deerfull}
\end{table}

The $\lambda=0.95$ base rows are the operating points reported in Table~\ref{tab:frontier} and discussed in Section~\ref{sec:comparison}.

\subsection{Per-seed detail}\label{app:perseed}
Section~\ref{sec:seeds} summarizes seed robustness; the full per-seed train-24 and AMC numbers are in Table~\ref{tab:perseed}, against a train-24 base of 0.812 / 6516 / 0.95 and an AMC base of 0.813 / 6280 / 0.89.

\begin{table}[t]
\centering
\small
\begin{tabular}{lrrrr}
\toprule
 & seed 17 & seed 0 & seed 42 & mean $\pm$ sd \\
\midrule
\multicolumn{5}{l}{\emph{train-24}} \\
acc & 0.792 & 0.771 & 0.823 & 0.795 $\pm$ 0.026 \\
think & 4911 & 5021 & 4724 & 4885 $\pm$ 150 \\
$\Delta$think & $-$25\% & $-$23\% & $-$28\% & $-$25.0\% $\pm$ 2.3 \\
closed & 0.99 & 0.99 & 1.00 & 0.99 \\
\addlinespace
\multicolumn{5}{l}{\emph{AMC}} \\
acc & 0.822 & 0.818 & 0.825 & 0.822 $\pm$ 0.004 \\
think & 4760 & 4719 & 4632 & 4704 $\pm$ 65 \\
$\Delta$think & $-$24\% & $-$25\% & $-$26\% & $-$25.1\% $\pm$ 1.0 \\
closed & 0.98 & 0.98 & 0.97 & 0.98 \\
\bottomrule
\end{tabular}
\caption{\textbf{Per-seed detail.} Full train-24 and AMC numbers for the three seeds; seed 17 is the headline adapter.}
\label{tab:perseed}
\end{table}

Closure stays at or above 0.97 for every seed. The per-benchmark base-against-adapter closure numbers across all five held-out benchmarks are in Table~\ref{tab:termination}.

\subsection{Full steering, termination, and burst-merge tables}\label{app:relocated}
The main text summarizes these results in prose and figures; the full tables are collected here. Table~\ref{tab:layersweep} is the four-layer steering sweep and Table~\ref{tab:layer6} the layer-6 sweep behind Section~\ref{sec:causal}, Table~\ref{tab:termination} the per-benchmark termination numbers behind Figure~\ref{fig:closure}, and Table~\ref{tab:burst} the burst-merge numbers behind Section~\ref{sec:scaling}.

\begin{table}[t]
\centering
\small
\begin{tabular}{rrll}
\toprule
layer & $\rho$(strength, length) & $+$10: think / term / acc & $+$25: think / term / acc \\
\midrule
6 & $-$0.63 & 981 / 77\% / 14\% & 27 / 67\% / 0\% \\
12 & $-$0.44 & 3685 / 75\% / 28\% & 8192 / 45\% / 2\% \\
\textbf{18} & \textbf{$-$0.82} & \textbf{3256 / 83\% / 56\%} & 18 / 89\% / 12\% \\
24 & $-$0.24 & 4619 / 78\% / 72\% & 3180 / 80\% / 61\% \\
\bottomrule
\end{tabular}
\caption{\textbf{Layer sweep for halt-vector steering.} Strength-to-length correlation $\rho$ and the two positive settings ($+$10, $+$25) at each layer (64 problems, strength $-$50 to $+$50; base at strength 0: 4{,}746 median think-tokens, 80\% terminated, 72\% accuracy). Layer 18 gives the strongest, cleanest length control.}
\label{tab:layersweep}
\end{table}

\begin{table}[t]
\centering
\small
\begin{tabular}{rrrrl}
\toprule
$\alpha$ (layer 6) & median think-tok & terminated & accuracy & value-axis control \\
\midrule
$-$50 & 12288 (cap) & 0\% & 1.6\% & flat \\
0 & 5188 & 88\% & 81\% & identical at $\alpha = 0$ \\
$+$25 & 32 & 68\% & 1.6\% & flat \\
$+$50 & 17 & 77\% & 0\% & flat \\
\bottomrule
\end{tabular}
\caption{\textbf{Steering the halt vector at layer 6.} Median think-tokens, termination rate, and accuracy as the steering strength $\alpha$ is swept (128 goldilocks problems, layer 6, $t=0.6$); a magnitude-matched value-axis control stays flat.}
\label{tab:layer6}
\end{table}

\begin{table}[t]
\centering
\small
\begin{tabular}{lrrrr}
\toprule
benchmark & base closed & ours closed & base non-term & ours non-term \\
\midrule
MATH500 & 0.97 & 0.99 & 3\% & 1\% \\
train-24 (goldilocks) & 0.95 & 0.99 & 5\% & 1\% \\
goldilocks holdout & 0.93 & 0.98 & 7\% & 2\% \\
AMC & 0.89 & 0.98 & 11\% & 2\% \\
AIME24 & 0.69 & 0.94 & 31\% & 6\% \\
AIME25 & 0.62 & 0.91 & 38\% & 9\% \\
\bottomrule
\end{tabular}
\caption{\textbf{Termination by difficulty.} Fraction emitting \texttt{</think>} (closed) and non-termination rate for the base model and reconstruction.}
\label{tab:termination}
\end{table}

\begin{table}[t]
\centering
\small
\begin{tabular}{rrrrr}
\toprule
bursts & problems & holdout $\Delta$think (acc) & AMC $\Delta$think (acc) & closed \\
\midrule
1 & 24 & $-$22.0\% (0.819) & $-$24.2\% (0.822) & 0.98 \\
2 & 48 & $-$24.2\% (0.811) & $-$26.6\% (0.809) & 0.98 \\
3 & 72 & $-$24.3\% (0.824) & $-$25.8\% (0.839) & 0.98 \\
4 & 96 & $-$23.8\% (0.816) & $-$26.1\% (0.809) & 0.98 \\
5 & 120 & $-$24.7\% (0.805) & $-$23.8\% (0.810) & 0.98 \\
6 & 144 & $-$24.9\% (0.825) & $-$26.2\% (0.815) & 0.98 \\
\bottomrule
\end{tabular}
\caption{\textbf{Burst-merge scaling.} Averaging independently-trained 24-problem LoRA bursts in weight space. Think reduction (accuracy in parentheses) holds between 24 and 26\% out to 144 problems, far from the eroded single run.}
\label{tab:burst}
\end{table}

\end{document}